\pdfoutput=1

\documentclass[11pt]{article}

\usepackage[]{emnlp2021}

\usepackage{times}
\usepackage{latexsym}

\usepackage[T1]{fontenc}

\usepackage[utf8]{inputenc}

\usepackage{microtype}
\usepackage{booktabs}
\usepackage{multirow}
\usepackage{adjustbox}
\usepackage{tabularx}
\usepackage{array}
\usepackage{graphicx,subcaption}
\usepackage{capt-of}
\usepackage{amsmath}
\usepackage{bm}

\title{End-to-End Cross-Domain Text-to-SQL \\ Semantic Parsing with Auxiliary Task}

\author{Peng Shi\textsuperscript{\rm 1}\thanks{\hspace{0.1cm}~Work done while at AWS AI Labs.}~,
  Tao Yu\textsuperscript{\rm 2},
  Patrick Ng\textsuperscript{\rm 3}, and Zhiguo Wang\textsuperscript{\rm 3} \\
\textsuperscript{\rm 1} David R. Cheriton School of Computer Science, University of Waterloo\\
\textsuperscript{\rm 2} Department of Computer Science, The University of Hong Kong \\
\textsuperscript{\rm 3} AWS AI Labs \\
{\tt peng.shi@uwaterloo.ca,
 \tt \{patricng,zhiguow\}@amazon.com} 
}

\date{}

\begin{document}
\maketitle
\begin{abstract}

In this work, we focus on two crucial components in the cross-domain text-to-SQL semantic parsing task:
\textit{schema linking} and \textit{value filling}. 
To encourage the model to learn better encoding ability, we propose a column selection auxiliary task to 
empower the encoder with the relevance matching capability by using explicit learning targets.
Furthermore, we propose two value filling methods to build the bridge from the existing zero-shot semantic parsers to real-world applications, considering most of the existing parsers ignore the \textit{values filling} in the synthesized SQL.
With experiments on Spider, our proposed framework improves over the baselines on the execution accuracy and exact set match accuracy when database contents are unavailable, and detailed analysis sheds light on future work.

\end{abstract}

\section{Introduction}

Text-to-SQL semantic parsing is to translate natural language questions into SQL queries \cite{warren1982efficient,popescu2003towards,li2014constructing,iyer17,Yu18,dong18}, 
which can be further executed against large scale databases to obtain query results.
These semantic parsers not only offer the chance of communicating with databases to the users who are unknown of the underlying database query language,
but are handy toolkits for experts to synthesize complex SQL queries by providing high-quality samples.

More recently, there has been significant interest in cross-domain text-to-SQL modeling innovation~\cite{Yu&al.18.emnlp.syntax,guo2019towards, Bogin2019RepresentingSS, wang2019rat,shi2020learning,yu2020grappa}, and different benchmarks are proposed, such as WikiSQL~\cite{zhong2017seq2sql} and Spider~\cite{yu2018spider}.
Especially for the Spider dataset, some of the SQL queries are nested which are more complicated than those in WikiSQL.
Furthermore, the attribute that training and test set do not have overlap databases makes the task zero-shot.
Therefore, linking entities in the question to the correct schema~(columns and tables) from the unseen databases becomes a major problem in the cross-domain text-to-SQL parsing task.
Another tricky problem is that some questions contain database-dependent values in different domains, which requires the model to extract them and fill in the corresponding positions for composing an executable SQL query.
However, this value filling problem is ignored by a lot of previous work in the cross-domain text-to-SQL parsing, making those parsers impractical.

\begin{figure}[t]
\centering{
  \includegraphics[width=0.45\textwidth]{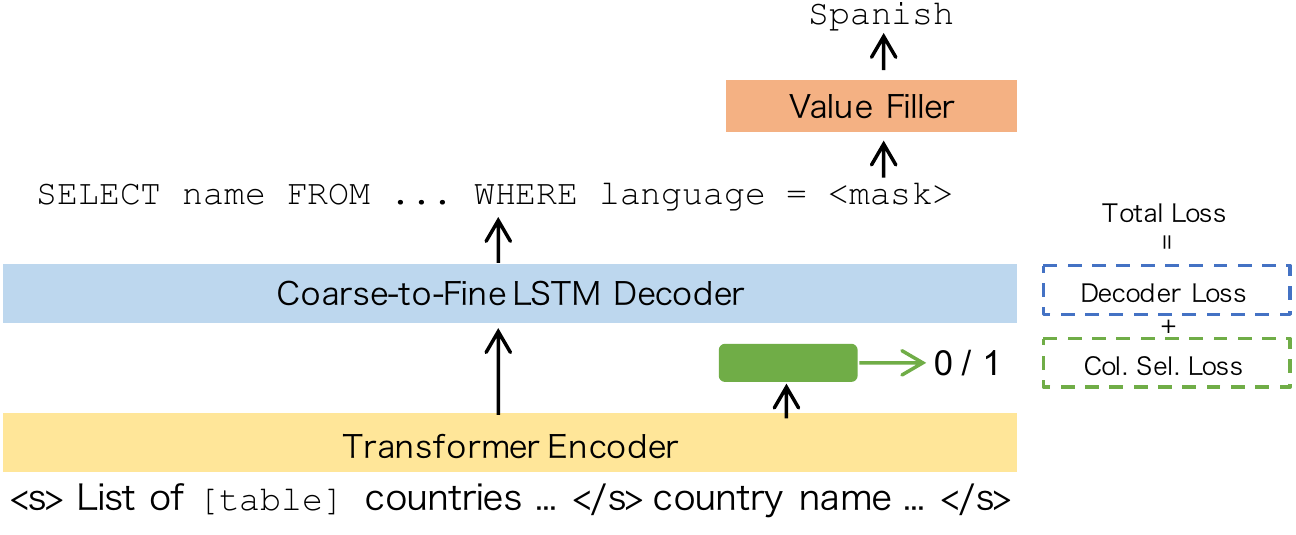}
  \caption{The overall architecture of our framework. The question is ``\textit{List of countries where Spanish is an official language.}"
  The encoder-decoder model firstly interprets the question into a SQL query with value placeholder \texttt{<mask>}. A value filler then chooses the best value for each placeholder to generate an executable SQL.
  The encoder-decoder model and the auxiliary task \textit{column selection} is trained jointly.}
  \label{fig:arch}}

  \vspace{-5mm}
\end{figure}

Considering the importance of the schema linking and value filling for a \textit{real-world} text-to-SQL parser in the cross-domain setting, we propose a framework with enhanced schema encoding ability and value filling ability.
In this framework, based on sequence-to-sequence architecture, we design an auxiliary task \textit{column classification} for multitask training.
The classifier determines whether a specific column is used in the gold SQL query, as shown in Figure~\ref{fig:arch}.
With this auxiliary task, the encoder has better relevance matching capability~(between the question and schema), supervised by explicit learning signals.
In this way, the decoder can distinguish relevant and irrelevant schema more easily, leading to better schema linking performance.
Furthermore, in our framework, we propose two value filling methods, including a heuristic method and a neural baseline, to make the zero-shot parsers usable in real-world applications.
With further analysis based on these two simple baselines, we provide more insights on the value filling subtask, paving ways for future work.

\section{Framework}

Given a question $X$ = $\{x_1, x_2, ..., x_n\}$ and schema $S$ consists of tables $T = \{t_1, t_2, ..., t_{|T|}\}$ and columns $C = \{c_1, c_2, ..., c_{|C|}\}$, our goal is to map the question-schema pair $(X, S)$ into the correct SQL query $Y$, which can be represented as a sequence of tokens of SQL, or other intermediate representations.
In this work, we adopt the intermediate representation as our learning target.

\subsection{Input, Encoding and Decoding}

\label{joint_encoding}
For a question, following \citet{yu2018typesql} and \citet{guo2019towards}, we firstly group tokens into a segment if that subsequence matches any column name or table name, and add a special indicator \texttt{[column]} or \texttt{[table]} to it, as shown in Figure~\ref{fig:arch}.
For the schema, we combine the column names with the corresponding table names. For example, table \textit{country} has a column \textit{population}, we then concatenate them into \textit{country population} as an enhanced column name.
In this way, the model can obtain better representation for distinguishing those columns that share the same column name but belong to different tables. 
Though Graph Neural Network \cite{bogin2019representing} or Relational Transformer \cite{wang2019rat} can achieve similar purpose using schema graph structure, we argue that this technique is simple to apply without introducing extra model complexity.

To encode the processed question and schema, we use the pretrained transformer BART~\cite{lewis2019bart}\footnote{We only leverage the encoder part of BART for the encoding, to save computational resources.} as the backbone.
The input sequence is the concatenation of processed question tokens and processed columns, as shown in Figure~\ref{fig:arch}.
After feeding the input sequence into the encoder, we then obtain the question representation $\{h_{x_1}, h_{x_2}, ..., h_{x_m}\}$ and column representation $\{ h_{c_1} \}$, $\{ h_{c_2} \}$, ..., $\{h_{c_{|C|}}\}$, 
where $\{h_{c_j}\}$ denotes a sequence of hidden vectors of $j-$th column, because each column contains several tokens.
We further apply two aggregation functions $f_{avg}$ and $f_{lstm}$ on the question and column representation, respectively. 
Concretely, the aggregation function $f_{avg}$ is an averaging function $\frac{1}{r-l+1} \sum_{i=l}^{r} h_{x_i}$ applying on each question segment which is determined in the pre-processing step.
The obtained question segment representation is denoted with $q_i$, which is further used for the decoding process.
The aggregation function $f_{lstm}$ is parameterized with a bidirectional LSTM and the last hidden states from both directions are concatenated to obtain the aggregated representation of each column.

For the decoder, we adopt the coarse-to-fine decoder from \citet{dong2018coarse} and \citet{guo2019towards}.
The decoder firstly generates the overall structure of the target SQL such as \texttt{SELECT \_\_ WHERE \_\_ ORDER BY \_\_}, 
and then conducts schema grounding in the fine-grained decoding process, including the \textit{column linking} and \textit{table linking}.
For more details about the coarse-to-fine deocder, please refer to \citet{dong2018coarse} and \citet{guo2019towards}.

\subsection{Multitask Learning}
To improve the schema encoding quality, an auxiliary \textit{column classification} task is designed for the model.
This auxiliary task exploits the direct supervision signals to enhance the encoder's relevance matching capability.
Specifically, we apply the aggregation function $f_{avg}$ on the representations of token sequence $\{h_{c_i}\}$ for each column, obtaining $s_{i}$.
An dot-product based attention mechanism is further applied on $s_{i}$ and question segment representations $\{q_j\}$, 
to obtain question-aware column representation $\hat{s}_{i}$ with equation $\hat{s}_{i} = \text{MLP}([s_{i}; r_{i}])$, where $r_{i} = \sum_{j=1}^m \alpha_j q_{j}$ and $\alpha_{j} \propto \text{exp}(s_{i}^{\top} q_{j})$.
Finally, a binary classifier is applied on $\hat{s}_i$: If the column is used in the target SQL sequence, then a positive label is assigned; Otherwise, it is classified with a negative label.

\subsection{Value Filler}

To bridge the gap between existing zero-shot semantic parsers~\cite{guo2019towards,bogin2019global,wang2019rat} and real-world applications, we propose two value fillers. 

\smallskip \noindent \textbf{Heuristic Method.} 
\label{heuristic}
Our heuristic method is based on the database search and string matching.
For each question token $x_i$, we search for its candidate cell values by executing 
the following SQL query

\noindent\fbox{\begin{minipage}{0.45\textwidth}
\fontfamily{Times}\selectfont
    \small
    \texttt{SELECT \{column\} FROM \{table\} WHERE \{column\} LIKE `\{token\} \%' OR \{column\} LIKE `\% \{token\}' OR \{column\} LIKE `\% \{token\} \%' OR \{column\} LIKE `\{token\}'}
   
\end{minipage}} 

\noindent for each column and the corresponding table. 
If there is a cell value string $v$ from the database is retrieved, and string $v$ has high similarity with a substring of the question~(measured by the Levenshtein distance), then we build a projection $P$ from \{table, column\} $\rightarrow$ \{$v$\}.  
We also collect all numbers mentioned in the question as number candidate list $N$, where the rule-based method is applied.
With the projection $P$ and number candidates $N$, we fill in the blank\footnote{We refer those positions in SQL that require values as \textit{blanks}.} with value $P$\{table, column\}, where the value is used in the specific table and column.
If multiple candidates are available, we choose the unused one in the queue, whose ordering is based on the collection sequence.
If the column type is \textit{number}, then a number will be assigned from the number candidate list $N$ based on the collection sequence.
If all numbers are used, the default value 1 is used.
If there is no projection for \{table, column\}, a placeholder is assigned.

\smallskip \noindent \textbf{Neural Baseline.}
Based on the value candidates including retrieved cell values in $P$ and the numbers in $N$, instead of filling the blanks with heuristic rules, we leverage the neural model for composing the executable SQL.
To achieve this, we firstly use a placeholder \texttt{<mask>}, as shown in Figure~\ref{fig:arch}, to denote the value to be filled in the incomplete SQL.
The input sequence for the model is the concatenation of the question, the incomplete SQL and value candidates, which is further fed into \texttt{roberta-base}~\cite{liu2019roberta} encoder, obtaining the \texttt{<mask>} token hidden states $u_{mask_{i}}$.
For the value candidates, averaging aggregation function $g_{avg}$ is applied to produce the value candidate representation $u_{v_j}$. 
By querying the value candidates with $u_{mask_{i}}$, best value candidate can be obtained based on $p(u_{v_j}|u_{mask_{i}}) \propto \text{exp}(u_{mask_{i}}^{\top} u_{v_j})$.

\section{Experimental Setup}

We conduct our experiments on the Spider~\cite{yu2018spider} dataset\footnote{Test set is not publicly available and model submission is required for obtaining test performance.}.
For the Spider dataset, there are two methods of evaluation: 
One is exact set match accuracy, requiring an exact set match between the predicted SQL and the oracle SQL, except for the values in the SQL; 
the other one is execution accuracy, requiring an exact match on the execution outputs.
In our work, we provide results on both evaluation metrics to provide a complete baseline system for the benchmark.
Considering some privacy issues in real world cases, database contents might be unavailable in the model development phase.
So in this work, for exact set match accuracy metric, we further evaluate our models under two different settings:
\textit{without using cell value} and \textit{using cell value} with \textit{one} model without retraining.
To achieve this, our model is trained without leveraging cell values from the database and can be evaluated directly when database contents are unavailable.
When the model can access the database contents, if a question token $x_i$ matches the cell values in column $c$, column name of $c$ is inserted before $x_i$, as a signal for the model.
More training details are provided in Appendix.

\section{Results and Analysis}

\begin{table}[t]
	\small
	\centering{
		\begin{tabular}{lcc}
			\toprule
			Model & Dev & Test \\
			\midrule 
			GAZP~(\citet{zhong2020grounded}) & - & 53.5 \\
			BRIDGE~(\citet{lin2020bridging}) & - & 59.9 \\ 
			\hline
		    Ours~(heuristic method) & 67.6 & \textbf{62.6} \\
			\bottomrule
	\end{tabular}}
	\caption{Results on execution accuracy in test set.}
	\label{table:exec-acc}
    \vspace{-3mm}
\end{table}

\begin{table}[t]
	\small
	\centering{
		\begin{tabular}{lcc}
			\toprule
			Model & Dev & Test \\
			\midrule 
			\multicolumn{3}{l}{\textbf{\textit{without using cell value}}} \\
			EditSQL~\cite{zhang19} v1 & 57.6 & 53.4 \\
			IRNet v2~\cite{guo2019towards} & 63.9 & 55.0 \\
			RASQL~(anonymous) & 60.8 & 55.7 \\
			RYANSQL v2~\cite{Choi2020RYANSQLRA} & 70.6 & 60.6 \\ 
			\hline
			Ours & 68.0 & \textbf{61.3} \\
			\hline\hline
			\multicolumn{3}{l}{\textbf{\textit{using cell value}}} \\
			BRIDGE~(\citet{lin2020bridging}) & 65.5 & 59.2 \\
			RATSQL v2~\cite{wang2019rat} & 65.8 & 61.9 \\
			RATSQL v3~\cite{wang2019rat} & 69.7 & \textbf{65.6} \\ 
			\hline
		    Ours & 70.0 & 61.9 \\
			\bottomrule
	\end{tabular}}
	\caption{Results on exact set match accuracy in test set~(we are comparing with leaderboard results on the submission date -- May 31, 2020).}
	\label{table:exact-acc}
	\vspace{-4mm}
\end{table}

For the end-to-end evaluation with execution accuracy, Table~\ref{table:exec-acc} shows the main results.
We report the results based on the heuristic value filler because we find that there is no significant difference between heuristic method and neural baseline model while the heuristic method is more efficient.
On the development set, our model achieves 67.6\% execution accuracy and 62.6\% on the test set.

We further show the results of exact set match accuracy in Table~\ref{table:exact-acc}, under two evaluation settings.
When the cell values are not available, our model achieves 68.0\% and 61.3\% accuracy on the development set and the hidden test set, respectively.
Note that RATSQL v2/v3 has significant performance drop~($>$7\%) when database contents are unavailable.
In comparison, our model is more robust under this setting.
When the database contents are available, with the simple method,
our model obtains 61.9\% exact set match accuracy on the hidden test set, which is comparable with RATSQL v2 with smaller model size~(RATSQL v2 leverages 24-layer pretrained transformer and 8-layer relational transformer with hidden size 1024 vs. Ours leverages 12-layer pretrained transformer with hidden size 1024).

\begin{table}[t]
	\small
	\centering{
		\begin{tabular}{lc}
			\toprule
			Model & Dev \\
			\midrule 
			\multicolumn{2}{l}{\textbf{\textit{exact set match accuracy}}} \\
			Full Model & 70.0 \\
			\rule{15pt}{0pt} - w/o Auxiliary Task & 68.0 \\
			\midrule
			\multicolumn{2}{l}{\textbf{\textit{execution accuracy}}} \\
			No Value Filler & 41.4 \\
			Heuristic Method & 67.6 \\
			Neural Baseline & 67.4 \\
			\bottomrule
	\end{tabular}}
	\caption{Ablation study on dev set.}
	\label{table:ablation}
	\vspace{-3mm}
\end{table}

\begin{table}[t]
	\small
	\centering{
		\begin{tabular}{lccccc}
			\toprule
			Split & Easy & Medium & Hard & Extra Hard & All \\
			\midrule 
			\multicolumn{6}{l}{\textbf{\textit{exact set match accuracy without using cell value}}} \\
			Dev & 87.4 & 72.6 & 54.0 & 42.7 & 68.0 \\
			Test & 81.9 & 67.7 & 52.5 & 30.3 & 61.3 \\
			\midrule 
			\multicolumn{6}{l}{\textbf{\textit{exact set match accuracy with using cell value}}} \\
			Dev & 87.8 & 74.7 & 58.0 & 44.5 & 70.0 \\
			Test & 83.0 & 68.8 & 51.8 & 30.8 & 61.9 \\
			\midrule 
			\multicolumn{6}{l}{\textbf{\textit{execution accuracy}}} \\
			Dev & 85.0 & 70.2 & 60.9 & 43.3 & 67.9 \\
			Test & 83.6 & 64.2 & 55.1 & 40.9 & 62.6 \\
			\bottomrule
	\end{tabular}}
	\caption{Accuracy breakdown for different hardness.}
	\label{table:exec-breakdown}
	\vspace{-4mm}
\end{table}

\smallskip \noindent \textbf{Effects of Auxiliary Task:}
As shown in first part of Table~\ref{table:ablation}, removing the auxiliary task leads to 2.0\% drop in exact set match accuracy. 
Further analysis shows that our full model has better effectiveness on easy and medium level queries, by improving easy-level query accuracy from 80.0\% to 87.8\%, and from 72.6\% to 74.7\% for medium-level query.

\smallskip \noindent \textbf{Performance of Value Fillers:}
Here, we are comparing the performance of the two value fillers, based on the predicted SQL by our full model. 
The results are shown in second part of Table~\ref{table:ablation}.
Here, we also provide \textit{no value filler} results as a baseline, where the predicted SQL from the parser is directly used for evaluation,
which obtains 41.4\% execution accuracy.
With the heuristic method, the model achieves 67.6\% on the execution accuracy, which is comparable with the neural baseline model~(67.4\%), 
showing that our heuristic method is effective as a baseline model in the value filling sub-task, and much simpler.
Based on error analysis, the model can be further improved if the model has the ability to link the values mentioned in question to actual values in database which have different surface forms, for example, \textit{female} and \textit{United States} in question should be linked to \textit{F}~(denotes female) and \textit{USA} in database.

\smallskip \noindent \textbf{Generalization}
As we can see from Table~\ref{table:exec-acc} and Table~\ref{table:exact-acc}, there is a large performance gap between the development set and hidden test set.
To investigate more on this issue, a breakdown of the accuracy by the query hardness level is provided in Table~\ref{table:exec-breakdown}, in all three settings.

From the results, we observe that for the exact set match accuracy, the gap between the development set and the test set on extra hard questions is large~(12.4\% and 13.7\%), which is aligned with the finding of \citet{wang2019rat}. 
However, another interesting finding is that the gap is much smaller on the execution accuracy, which is only 2.4\% drop.
The reason for this remains unknown because the test set is not available.
However, one possible reason is that our model provides semantically equivalent predictions as the oracle SQL, 
leading to the same execution outputs,
while the SQL predictions fail in the exact set match testing which requires syntactical equivalence.
For example, the keywords \texttt{EXCEPT} and \texttt{NOT IN} can be used to express same meaning, and similar for \texttt{INTERSECT} and \texttt{AND} in \texttt{WHERE} conditions.
Based on this observation, we suggest reporting both the exact set match accuracy and execution accuracy in the future work, to gain better understanding of the proposed methods.

\section{Conclusions}
In this work, we leverage an auxiliary task to enhance the schema encoding ability of encoder.
We also propose two value filling baseline models to bridge the gap between the existing models and a usable text-to-SQL parser.
With experiments on Spider dataset, we show improved performance over the baselines, with the execution accuracy and exact set match accuracy when database contents are unavailable.

\bibliographystyle{acl_natbib}
\bibliography{custom}

\begin{thebibliography}{24}
\expandafter\ifx\csname natexlab\endcsname\relax\def\natexlab#1{#1}\fi

\bibitem[{Bogin et~al.(2019{\natexlab{a}})Bogin, Berant, and
  Gardner}]{Bogin2019RepresentingSS}
Ben Bogin, Jonathan Berant, and Matt Gardner. 2019{\natexlab{a}}.
\newblock Representing schema structure with graph neural networks for
  text-to-sql parsing.
\newblock In \emph{ACL}.

\bibitem[{Bogin et~al.(2019{\natexlab{b}})Bogin, Gardner, and
  Berant}]{bogin2019global}
Ben Bogin, Matt Gardner, and Jonathan Berant. 2019{\natexlab{b}}.
\newblock Global reasoning over database structures for text-to-sql parsing.
\newblock \emph{arXiv preprint arXiv:1908.11214}.

\bibitem[{Bogin et~al.(2019{\natexlab{c}})Bogin, Gardner, and
  Berant}]{bogin2019representing}
Ben Bogin, Matt Gardner, and Jonathan Berant. 2019{\natexlab{c}}.
\newblock Representing schema structure with graph neural networks for
  text-to-sql parsing.
\newblock \emph{arXiv preprint arXiv:1905.06241}.

\bibitem[{Choi et~al.(2020)Choi, Shin, Kim, and Shin}]{Choi2020RYANSQLRA}
Donghyun Choi, Myeong~Cheol Shin, Eunggyun Kim, and Dong~Ryeol Shin. 2020.
\newblock Ryansql: Recursively applying sketch-based slot fillings for complex
  text-to-sql in cross-domain databases.
\newblock \emph{ArXiv}, abs/2004.03125.

\bibitem[{Dong and Lapata(2018{\natexlab{a}})}]{dong18}
Li~Dong and Mirella Lapata. 2018{\natexlab{a}}.
\newblock Coarse-to-fine decoding for neural semantic parsing.
\newblock In \emph{Proceedings of the 56th Annual Meeting of the Association
  for Computational Linguistics (Volume 1: Long Papers)}, pages 731--742.
  Association for Computational Linguistics.

\bibitem[{Dong and Lapata(2018{\natexlab{b}})}]{dong2018coarse}
Li~Dong and Mirella Lapata. 2018{\natexlab{b}}.
\newblock Coarse-to-fine decoding for neural semantic parsing.
\newblock \emph{arXiv preprint arXiv:1805.04793}.

\bibitem[{Guo et~al.(2019)Guo, Zhan, Gao, Xiao, Lou, Liu, and
  Zhang}]{guo2019towards}
Jiaqi Guo, Zecheng Zhan, Yan Gao, Yan Xiao, Jian-Guang Lou, Ting Liu, and
  Dongmei Zhang. 2019.
\newblock Towards complex text-to-sql in cross-domain database with
  intermediate representation.
\newblock \emph{arXiv preprint arXiv:1905.08205}.

\bibitem[{Iyer et~al.(2017)Iyer, Konstas, Cheung, Krishnamurthy, and
  Zettlemoyer}]{iyer17}
Srinivasan Iyer, Ioannis Konstas, Alvin Cheung, Jayant Krishnamurthy, and Luke
  Zettlemoyer. 2017.
\newblock Learning a neural semantic parser from user feedback.
\newblock \emph{CoRR}, abs/1704.08760.

\bibitem[{Lewis et~al.(2019)Lewis, Liu, Goyal, Ghazvininejad, Mohamed, Levy,
  Stoyanov, and Zettlemoyer}]{lewis2019bart}
Mike Lewis, Yinhan Liu, Naman Goyal, Marjan Ghazvininejad, Abdelrahman Mohamed,
  Omer Levy, Ves Stoyanov, and Luke Zettlemoyer. 2019.
\newblock Bart: Denoising sequence-to-sequence pre-training for natural
  language generation, translation, and comprehension.
\newblock \emph{arXiv preprint arXiv:1910.13461}.

\bibitem[{Li and Jagadish(2014)}]{li2014constructing}
Fei Li and HV~Jagadish. 2014.
\newblock Constructing an interactive natural language interface for relational
  databases.
\newblock \emph{VLDB}.

\bibitem[{Lin et~al.(2020)Lin, Socher, and Xiong}]{lin2020bridging}
Xi~Victoria Lin, Richard Socher, and Caiming Xiong. 2020.
\newblock Bridging textual and tabular data for cross-domain text-to-sql
  semantic parsing.
\newblock \emph{arXiv preprint arXiv:2012.12627}.

\bibitem[{Liu et~al.(2019)Liu, Ott, Goyal, Du, Joshi, Chen, Levy, Lewis,
  Zettlemoyer, and Stoyanov}]{liu2019roberta}
Yinhan Liu, Myle Ott, Naman Goyal, Jingfei Du, Mandar Joshi, Danqi Chen, Omer
  Levy, Mike Lewis, Luke Zettlemoyer, and Veselin Stoyanov. 2019.
\newblock Roberta: A robustly optimized bert pretraining approach.
\newblock \emph{arXiv preprint arXiv:1907.11692}.

\bibitem[{Popescu et~al.(2003)Popescu, Etzioni, and Kautz}]{popescu2003towards}
Ana-Maria Popescu, Oren Etzioni, and Henry Kautz. 2003.
\newblock Towards a theory of natural language interfaces to databases.
\newblock In \emph{Proceedings of the 8th international conference on
  Intelligent user interfaces}, pages 149--157. ACM.

\bibitem[{Shi et~al.(2020)Shi, Ng, Wang, Zhu, Li, Wang, Santos, and
  Xiang}]{shi2020learning}
Peng Shi, Patrick Ng, Zhiguo Wang, Henghui Zhu, Alexander~Hanbo Li, Jun Wang,
  Cicero Nogueira~dos Santos, and Bing Xiang. 2020.
\newblock Learning contextual representations for semantic parsing with
  generation-augmented pre-training.
\newblock \emph{arXiv preprint arXiv:2012.10309}.

\bibitem[{Wang et~al.(2019)Wang, Shin, Liu, Polozov, and
  Richardson}]{wang2019rat}
Bailin Wang, Richard Shin, Xiaodong Liu, Oleksandr Polozov, and Matthew
  Richardson. 2019.
\newblock Rat-sql: Relation-aware schema encoding and linking for text-to-sql
  parsers.
\newblock \emph{arXiv preprint arXiv:1911.04942}.

\bibitem[{Warren and Pereira(1982)}]{warren1982efficient}
David~HD Warren and Fernando~CN Pereira. 1982.
\newblock An efficient easily adaptable system for interpreting natural
  language queries.
\newblock \emph{Computational Linguistics}, 8(3-4):110--122.

\bibitem[{Yu et~al.(2018{\natexlab{a}})Yu, Li, Zhang, Zhang, and Radev}]{Yu18}
Tao Yu, Zifan Li, Zilin Zhang, Rui Zhang, and Dragomir Radev.
  2018{\natexlab{a}}.
\newblock Typesql: Knowledge-based type-aware neural text-to-sql generation.
\newblock In \emph{Proceedings of NAACL}. Association for Computational
  Linguistics.

\bibitem[{Yu et~al.(2018{\natexlab{b}})Yu, Li, Zhang, Zhang, and
  Radev}]{yu2018typesql}
Tao Yu, Zifan Li, Zilin Zhang, Rui Zhang, and Dragomir Radev.
  2018{\natexlab{b}}.
\newblock Typesql: Knowledge-based type-aware neural text-to-sql generation.
\newblock \emph{arXiv preprint arXiv:1804.09769}.

\bibitem[{Yu et~al.(2020)Yu, Wu, Lin, Wang, Tan, Yang, Radev, Socher, and
  Xiong}]{yu2020grappa}
Tao Yu, Chien-Sheng Wu, Xi~Victoria Lin, Bailin Wang, Yi~Chern Tan, Xinyi Yang,
  Dragomir Radev, Richard Socher, and Caiming Xiong. 2020.
\newblock Grappa: Grammar-augmented pre-training for table semantic parsing.
\newblock \emph{arXiv preprint arXiv:2009.13845}.

\bibitem[{Yu et~al.(2018{\natexlab{c}})Yu, Yasunaga, Yang, Zhang, Wang, Li, and
  Radev}]{Yu&al.18.emnlp.syntax}
Tao Yu, Michihiro Yasunaga, Kai Yang, Rui Zhang, Dongxu Wang, Zifan Li, and
  Dragomir Radev. 2018{\natexlab{c}}.
\newblock Syntaxsqlnet: Syntax tree networks for complex and cross-domain
  text-to-sql task.
\newblock In \emph{Proceedings of EMNLP}. Association for Computational
  Linguistics.

\bibitem[{Yu et~al.(2018{\natexlab{d}})Yu, Zhang, Yang, Yasunaga, Wang, Li, Ma,
  Li, Yao, Roman et~al.}]{yu2018spider}
Tao Yu, Rui Zhang, Kai Yang, Michihiro Yasunaga, Dongxu Wang, Zifan Li, James
  Ma, Irene Li, Qingning Yao, Shanelle Roman, et~al. 2018{\natexlab{d}}.
\newblock Spider: A large-scale human-labeled dataset for complex and
  cross-domain semantic parsing and text-to-sql task.
\newblock \emph{arXiv preprint arXiv:1809.08887}.

\bibitem[{Zhang et~al.(2019)Zhang, Yu, Er, Shim, Xue, Lin, Shi, Xiong, Socher,
  and Radev}]{zhang19}
Rui Zhang, Tao Yu, He~Yang Er, Sungrok Shim, Eric Xue, Xi~Victoria Lin, Tianze
  Shi, Caiming Xiong, Richard Socher, and Dragomir Radev. 2019.
\newblock Editing-based sql query generation for cross-domain context-dependent
  questions.
\newblock In \emph{Proceedings of the 2019 Conference on Empirical Methods in
  Natural Language Processing and 9th International Joint Conference on Natural
  Language Processing}. Association for Computational Linguistics.

\bibitem[{Zhong et~al.(2020)Zhong, Lewis, Wang, and
  Zettlemoyer}]{zhong2020grounded}
Victor Zhong, Mike Lewis, Sida~I Wang, and Luke Zettlemoyer. 2020.
\newblock Grounded adaptation for zero-shot executable semantic parsing.
\newblock \emph{arXiv preprint arXiv:2009.07396}.

\bibitem[{Zhong et~al.(2017)Zhong, Xiong, and Socher}]{zhong2017seq2sql}
Victor Zhong, Caiming Xiong, and Richard Socher. 2017.
\newblock Seq2sql: Generating structured queries from natural language using
  reinforcement learning.
\newblock \emph{arXiv preprint arXiv:1709.00103}.

\end{thebibliography}

\newpage
\clearpage

\appendix

\section{Appendices}

\subsection{Experimental Setup}
Within the encoder, the hidden size is 1024 for the pretrained transformer.
For the LSTM in the aggregation function $f_{lstm}$, the hidden size is 512.
The hidden size of parameters in the decoder LSTM and the binary classifier of auxiliary task is 300.
The overall model size is 822M in disk.
Adam optimizer with default hyper-parameters is used. 
Batch size is 32 and accumulating gradient is applied.
The learning rate is set to 5e-4 for the non-pretrained parameters and 2.5e-5 for pretrained parameters.
The weight between the main parsing task and the auxiliary task is 1:0.5.
The experiment is conducted with Tesla V100 16GB and PyTorch version 1.3.0.
The Levenshtein distance is calculated by the FuzzyWuzzy~\footnote{\url{https://github.com/seatgeek/fuzzywuzzy}} library.
The exact set match accuracy and execution accuracy is calculated by the official evaluation script~\footnote{\url{https://github.com/taoyds/spider/blob/master/evaluation.py}}.

The model is trained up to 20K steps for 10 hours on single GPU.
The best parsing model is selected based on the exact set match accuracy on development set.
For the neural value filler, query level accuracy is used for selecting best model.

\end{document}